\documentclass[10pt,twocolumn,letterpaper]{article}
\usepackage[pagenumbers]{cvpr}      

\usepackage{amsmath}
\usepackage{amssymb}
\usepackage{booktabs}

\usepackage[pdftex]{graphicx}
\usepackage[table]{xcolor}

\usepackage{graphicx}

\DeclareMathOperator*{\argmin}{arg\,min}
\usepackage{multirow}
\def\BibTeX{{\rm B\kern-.05em{\sc i\kern-.025em b}\kern-.08em
    T\kern-.1667em\lower.7ex\hbox{E}\kern-.125emX}}

\usepackage{booktabs}

\usepackage[pagebackref,breaklinks,colorlinks]{hyperref}

\usepackage[capitalize]{cleveref}
\crefname{section}{Sec.}{Secs.}
\Crefname{section}{Section}{Sections}
\Crefname{table}{Table}{Tables}
\crefname{table}{Tab.}{Tabs.}

\def\cvprPaperID{*****} 
\def\confName{CVPR}
\def\confYear{2023}

\begin{document}

\title{LD-GAN: Low-Dimensional Generative Adversarial Network for Spectral Image Generation with Variance Regularization}

\author{Emmanuel Martinez, Roman Jacome, Alejandra Hernandez-Rojas and Henry Arguello\\
Universidad Industrial de Santander\\
{\tt\small \{emmanuel2162134@correo,roman2162474@correo,maria.hernandez26@correo,henarfu@\}.uis.edu.co}
\thanks{This paper was supported by the Vicerrectoría de Investigación y Extensión UIS, project code 3735}}

\maketitle

\begin{abstract}
Deep learning methods are state-of-the-art for spectral image (SI) computational tasks. However, these methods are constrained in their performance since {available} datasets are limited due to the highly expensive and long acquisition time. Usually, data augmentation techniques are employed to mitigate the lack of data. Surpassing classical augmentation methods, such as geometric transformations, GANs enable diverse augmentation by learning and sampling from the data distribution. Nevertheless, GAN-based SI generation is challenging since the high-dimensionality nature of this kind of data hinders the convergence of the GAN training yielding to suboptimal generation. To surmount this limitation, we propose low-dimensional GAN (LD-GAN), where we train the GAN employing a low-dimensional representation of the {dataset} with the latent space of a pretrained autoencoder network. Thus, we generate new low-dimensional samples which are then mapped to the SI dimension with the pretrained decoder network. Besides, we propose a statistical regularization to control the low-dimensional representation variance for the {autoencoder} training and to achieve high diversity of samples generated with the GAN. We validate our method {LD-GAN} as data augmentation {strategy} for compressive spectral imaging, SI super-resolution, and RBG to spectral tasks with improvements varying from 0.5 to 1 [dB] in each task respectively. {We perform comparisons} against the non-data augmentation training, traditional DA, and with {the same} GAN {adjusted} and trained to generate the full{-sized SIs}. {The code of this paper can be found in \url{https://github.com/romanjacome99/LD_GAN.git}}
\end{abstract}

\section{Introduction}
\label{sec:intro}

Spectral imaging involves acquiring {a collection of} 2D images {which contain specific electromagnetic radiation from light}, known as a spectral image (SI). The spectral information allows the estimation of unique characteristics and distribution of the different materials. SIs have been used for computer vision tasks related to object classification \cite{classification}, image segmentation \cite{segmentation}, salient object detection \cite{object_detection}, and object tracking \cite{object_tracking}, which have been widely applied in fields such as medical applications \cite{medical}, earth observation \cite{morales2023hyperspectral}, food quality \cite{food}, and surveillance \cite{survilleance}, among others.

The recent advances in data-driven deep learning (DL) methods have opened new frontiers for SI processing, acquisition, and its applications \cite{ozdemir2020deep}. Some examples are in hyperspectral, and multispectral image fusion \cite{fusiondl,jacome2021deep, jacome2022d}, classification \cite{classification, bacca2022deep_cod}, recovery methods for snapshot compressive spectral imaging (CSI) recovery \cite{huang2022spectral,bacca2023computational} or mapping RGB images to SI \cite{yan2018accurate}. While one of the main reasons for the great success of DL in a wide range of applications is that the models can extract the intrinsic structure of large datasets \cite{lecun2015deep} improving its generalization performance, in SI applications the datasets are limited in the number of available samples due to the expensive and long acquisition times \cite{snapshot}. Thus the performance of the DL methods is still restricted to the constrained available data. 

To address this issue, data augmentation (DA) strategies are employed \cite{shorten2019survey}. Traditional DA performs geometrical transformations, such as flipping, rotation, and {resizing} the original dataset to generate new training samples. However, these approaches generate few invariances on the original dataset, which are easily learned by the network. Recent approaches employ generative adversarial networks (GANs) \cite{goodfellow2020generative} to synthesize new samples based on learning the probability distribution of the original dataset and generating synthetic samples from the learned distribution \cite{antoniou2017data}. In the context of SI generation, {many works have been developed as a data augmentation strategy for SI classification~\cite{hang2020classification}, where conditional GANs are employed to generate new labeled data. A more related work is} SHS-GAN \cite{hauser2021shs}{, which} is a conditional GAN that creates new SI samples by mapping SIs from RGB images through a generator and employs a critic network to inject spectral information in the mapped SIs. Also, methods based on variational autoencoders (VAE) have been proposed to generate only spectral signatures in \cite{phillips2022variational}. {Unlike current SI generation methods that are  conditioned on labels or RGB images, we look to generate new SI samples from unconditional GANs leveraging the available spectral image datasets without additional information. }

{However}, training GAN for SI generation is challenging due to the high dimensionality and the limited training of SI samples. Theoretical studies describe these issues by deriving the convergence rate of the GAN estimator, yielding that for high dimensional data and a low number of samples, the convergence rate decreases, producing a sub-optimal performance of the sample generation \cite{huang2022error, liang2021well, arora2017generalization}. Experimentally, projected GAN \cite{sauer2021projected} has addressed this issue by training GANs using LD image datasets obtained from several feature or codification networks. Unlike the projected GAN, we look to develop a straightforward strategy for generating a unique LD representation of the SIs for training a GAN. {This approach differs from existing generative models based on dimensionality reduction such as  VAE\cite{kingma2013auto}. These methods aim to learn the dataset distribution via latent variables that represent the mean and variance of the distribution and with a Kullback-Leibler divergence regularization approximate this distribution to Gaussian.  Here, we propose to learn the latent distribution with a GAN and decode the image with a pre-trained decoder network.} {Additionally LD representations of SIs are very suitable since the high dimensionality of SIs produces high information redundancy in this type of data. Other SI applications have harnessed this in tasks such as the recovery of SIs from CSI measurements \cite{monroy2022jr2net}, classification \cite{alhayani2017hyper} or unmixing \cite{yi2017dimensionality}.} 

The proposed LD-GAN for generating SIs trains an autoencoder (AE) network using a SI dataset to obtain an LD image dataset. Then, a GAN is trained adversarially using the LD image dataset to generate new LD image samples. Finally, the generated LD image samples are decoded through the AE network to obtain generated SIs.  We propose a statistical regularization function over the LD representation images in the AE training. This regularization function promotes a variance minimization on LD space along the batch of the training data, thus having a more compact representation of the SI, yielding optimal decoding. This criterion follows from contraction AE that improves the {LD} representation by reducing the variability of the representation concerning the decoder's input \cite{rifai2011higher}. Additionally, since we are reducing the dimensionality of the generated data, 
in order to produce more diverse LD samples that results in more variability of the SI,  the generated LD images are regularized for the GAN training by maximizing the variance of the LD image.

{Since we are proposing an LD adversarial training method, we also develop a high dimensional adversarial training with the same GAN model in order to perform comparisons but adjusting the subnetworks to generate the full-sized SIs, which we will call from now on {S-GAN}.} Experimental results show that the proposed LD-GAN improves upon the convergence of the GAN compared with S-GAN. This improvement is further supported by the visual quality of the generated SIs, which show that the proposed method produces more realistic spatial and spectral information compared to the S-GAN. We validate the proposed LD-GAN as a DA technique for CSI recovery, SI spatial-spectral super-resolution, and RGB to spectral tasks. We present improvements varying from 0.5 to 1 [dB] with the proposed method compared with the non-DA training, traditional DA, and with a GAN trained to generate the full spectral image. 

The rest of the paper is organized as follows. Section \ref{sec:gan} is formulated as the traditional GAN. Thereafter, in section \ref{sec:ld_gan}, we formulate the proposed LD-GAN. Then, section \ref{sec:stats} describes the statistical regularization over the AE latent space and the GAN-generated LD samples. Section \ref{sec:applications} presents the mathematical formulation of the SI applications employed to validate the proposed methods. The experimental results are disseminated in section \ref{sec:results}. Finally, conclusions are presented in \ref{sec:conclusion}.

\section{Generative Adversarial Network}
\label{sec:gan}
\begin{figure*}[!t]
    \includegraphics[width=1\textwidth]{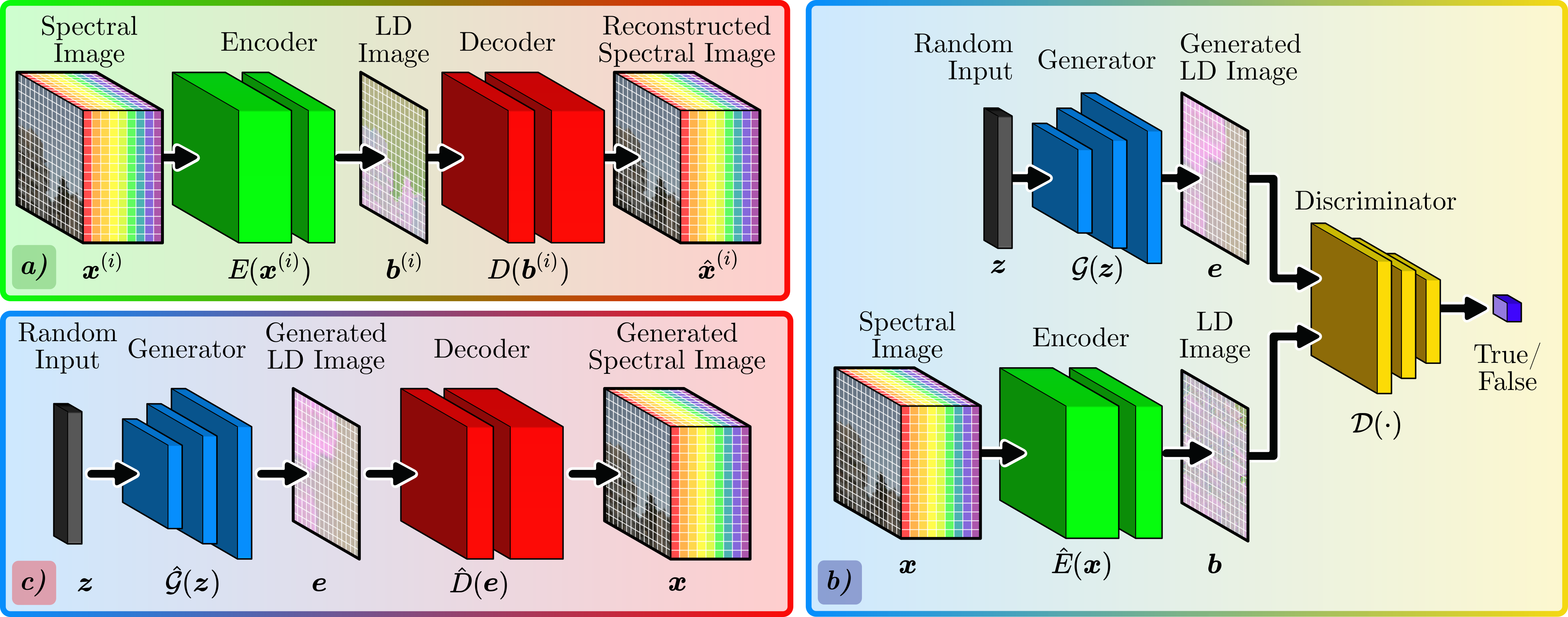}
    \centering
\vspace{-0.2cm}    \caption{{Autoencoder and LD-GAN networks.} a) The AE network {allows obtaining an LD representation of the SI} through an encoder subnetwork and then reconstructs the original SI with a decoder subnetwork. b) The LD representation obtained by the AE is employed to train a GAN to generate new LD representation images. c) Once the GAN is trained, the new LD images can be decoded to recover SIs.}
    \vspace{-5mm}
    \label{fig:autoencoder}
\end{figure*}

Given a dataset $\left\{\boldsymbol{x}^{(i)}\right\}_{i=1}^K$, with $K$ images and $\boldsymbol{x}^{(i)} \in \mathbb{R}^{MNL}$, where $M,N$ are the spatial dimensions and $L$ the number of spectral bands and assuming that the dataset follows a distribution $p_{\text {data}}(\boldsymbol{x})$, which is unknown. Then, a generative network $\mathcal{G}$ will learn the distribution $p_g$ from data $\boldsymbol{x}$. For the generative network, a prior is assumed on input noise variables $p_{\boldsymbol{z}}(\boldsymbol{z})$ where $\boldsymbol{z}\in \mathbb{R}^{m}$ and $m$ is the random variable dimension, usually a Gaussian distribution, which are mapped to the desired generated images. Then, a discriminative network $\mathcal{D}$ is also defined. This network will receive the generated samples or dataset samples. The loss function of a GAN \cite{goodfellow2020generative} is
\begin{align}
    \mathcal{L}_{gan}(\mathcal{D}, \mathcal{G}, \mathcal{I}) = \hspace{1mm} & \mathbb{E}_{\boldsymbol{x} \sim p_{\text {data}}(\boldsymbol{x})}\left[\log \left(\mathcal{D}(\mathcal{I}(\boldsymbol{x}))\right)\right]+\nonumber\\
    &\mathbb{E}_{\boldsymbol{z} \sim p_{\boldsymbol{z}}(\boldsymbol{z})}\left[\log \left(1-\mathcal{D}\left((\mathcal{G}(\boldsymbol{z})\right)\right)\right],
\end{align}
where $\mathcal{G}$ and $\mathcal{D}$ are the generator and distriminator networks, respectively. Additionally, $\mathcal{I}$ is an identity operator. In the adversarial training, each network $\{\mathcal{G}, \mathcal{D}\}$ compete{s} to achieve {its} goals: $\mathcal{G}$ will generate fake samples, and $\mathcal{D}$ will predict if the received samples are real or fake. This is called adversarial training and can be represented as
\begin{align}\label{eq:gan}
\{ \hat{\mathcal{G}}, \hat{\mathcal{D}} \} = \hspace{1mm} \underset{\mathcal{D}}{\arg \min} \hspace{1mm} \underset{\mathcal{G}}{\arg \max} \hspace{1mm} \mathcal{L}_{gan}(\mathcal{D}, \mathcal{G}, \mathcal{I}).
\end{align}
Despite GANs having achieved state-of-the-art results in the generation of images, achieving proper convergence of GAN through adversarial training is challenging when the employed dataset samples have high dimensionality and it is limited in the number of samples, such as the generation of SIs. In the following section, we will analyze the convergence of the GANs.
\subsection{Analysis on the GAN Convergence Rate}
Theoretical studies of the convergence properties of GAN have been carried out concerning the dimensionality of the data and the number of training samples. For instance, in \cite{liang2021well} derives an asymptotic bound of the total-variation metric between an estimated distribution and the real source data distribution, where the sample complexity relies on the squared dimension of the data (in our case $MNL$), and a logarithmic factor of the number of samples of the training dataset \cite[{Theorem} 19]{liang2021well}. Also, in \cite{huang2022error} provides an error analysis of GANs through convergence rates of the integral probability metric, yielding sample complexity that increases in terms of the number of training samples and the dimension of the data \cite[Theorem 5]{huang2022error}. Moreover, the convergence rate is improved when the source data can be represented in an LD manifold \cite[Theorem 19]{huang2022error}. Consequently, based on these theoretical insights on the GANs convergence, we propose generating an LD representation of the SIs to improve convergence on GANs and generate high-spectral dimensional SIs by decoding them through the pretrained AE network, described in the following section.
\section{Generative Adversarial Networks with Low-Dimensional Representation}
\label{sec:ld_gan}
This section will describe the proposed LD-GAN where LD samples are generated by a GAN and the synthetic SIs are obtained by decoding the LD via a decoder network. 
\subsection{Spectral Image Autoencoder}
An AE network is employed to obtain an LD representation of a SI dataset $\{\boldsymbol{x}^{(i)} \}_{i=1}^K$ with $K$ samples. Two subnetworks give the structure of an AE, as shown in Fig.~\ref{fig:autoencoder}(a). An encoder network $E$, which will compress the {spectral} information of SIs, and a decoder network $D$, which will decode the underlying SI. The optimization of these subnetworks is given by
\begin{equation}
\{\hat{E}, \hat{D} \} = \sum_{i=1}^K || \boldsymbol{x}^{(i)} - D(E(\boldsymbol{x}^{(i)}))||_2^2,
\end{equation}
where $\hat{E}$ and $\hat{D}$ are the encoder and decoder, respectively. Thus, an LD representation of a SI can be obtained as $\boldsymbol{b}^{(i)} = E(\boldsymbol{x}^{(i)})$ where $\boldsymbol{b}^{(i)} \in \mathbb{R}^{{MNc}}$ and ${c} < L$. Finally, SI can be recovered by decoding this LD image as $\hat{\boldsymbol{x}}^{(i)} = D(\boldsymbol{b}^{(i)})$.

\subsection{Low-Dimensional Generative Adversarial Network}
We {employ} a GAN that will be optimized with respect to the LD representation of {a} SI dataset. Then, the adversarial training from the equation \ref{eq:gan} can be rewritten as
\begin{align}\label{eq:proposed}
\{ \hat{\mathcal{G}}, \hat{\mathcal{D}} \} = \underset{\mathcal{D}}{\arg \min} \hspace{1mm} \underset{\mathcal{G}}{\arg \max} \hspace{1mm} \mathcal{L}_{gan}(\mathcal{D}, \mathcal{G}, \hat{E}),
\end{align}
where the generative network can generate LD images as $\boldsymbol{e} = \hat{\mathcal{G}}(\boldsymbol{z})$ and new SI samples can be obtained using the decoder as $\hat{\boldsymbol{f}} = \hat{D}(\boldsymbol{e})$.
\section{Statistical Regularization for the AE and GAN}
\label{sec:stats}
Towards improving the LD representations of the AE and the diversity of generated LD images by the GAN, we propose a variance minimization regularizer in the AE training that allows a compact representation of the SI dataset in the LD space, which improves AE recovery performance. Then, we employed a variance maximization on the generated LD space for the GAN training to produce diverse data and more quality on the {generated SI} dataset. This variance-based regularization criterion has also been used for self-supervised learning \cite{bardes2022vicreg} and sparse-coding \cite{evtimova2021sparse}.  First, define set $\mathcal{A} = \{\boldsymbol{e}^{(i)}\}_{i=1}^{K}$ that contains the {LD} representation of the AE and define the set of generated LD images {by GAN as} $\mathcal{W} =  \{\boldsymbol{d}^{(i)}\}_{i=1}^{K}$. The variance of $\mathcal{A}$ is denoted by $\boldsymbol{\sigma}_{\mathcal{A}}^{{2}}\in \mathbb{R}_+^d$, and of the GAN generated representations $\boldsymbol{\sigma}_{\mathcal{W}}^{{2}}\in \mathbb{R}_+^d$ pixel-wise across the training batch dimension. The proposed statistical regularization function is given by
\begin{equation}
    \mathcal{R}(\cdot) = \Vert \boldsymbol{\sigma}^{{2}}_{(\cdot)}\Vert_2.
\end{equation}
Then, the regularized AE training is performed as follows 
\begin{equation}
\{\hat{E}, \hat{D} \} = \underset{E, D}{\arg \min} \hspace{1mm} || \boldsymbol{x}^{(i)} - D(E(\boldsymbol{x}^{(i)}))||_2^2 + \mu_{ae}\mathcal{R}(\mathcal{A}).
\end{equation}
and the final GAN optimization problem is given by
\begin{align}\label{eq:proposed_reg}
\{ \hat{\mathcal{G}}, \hat{\mathcal{D}} \} = \underset{\mathcal{D}}{\arg \min} \hspace{1mm} \underset{\mathcal{G}}{\arg \max} \hspace{1mm}  \mathcal{L}_{gan}(\mathcal{D}, \mathcal{G}, \hat{E}) - \mu_{gan} \mathcal{R}(\mathcal{W}) .
\end{align}
Note that the negative sign in the variance regularization function is because we want to maximize the variance of the generated LD representation of the SI. In both optimization problems, the AE and the GAN, the parameters $\mu_{ae}$ and $\mu_{gan}$ are regularization hyperparameters that control how much we concentrate the AE {LD} space or how much we increase the variability of the generated images by the GAN.
\section{Spectral Image Applications}\label{sec:applications}
To validate the performance of our proposed method, we address the following DL-based SI tasks:

\textbf{CSI Recovery~\cite{bacca2023computational}:} CSI is a technique that allows sensing the whole 3D information of the SI with a 2D sensor in a single shot \cite{yuan2021snapshot} through coded projections of the high-dimensional data {and} a computational algorithm to recover the underlying SI. Several optical architectures have been proposed for CSI. We employed the coded aperture snapshot spectral imager (CASSI) \cite{CASSI}, which is the flagship architecture in CSI \cite{CSI}. This sensing phenomenon is expressed in a matrix-vector multiplication as $\boldsymbol{y} = \boldsymbol{H}\boldsymbol{x} + \boldsymbol{n}$, where $\boldsymbol{y} \in \mathbb{R}^{{M}({N}+L-1)}$ are the compressed measurements, $\boldsymbol{H}\in\mathbb{R}^{{M}({N}+L-1)\times MNL}$ is the CASSI sensing matrix, $\boldsymbol{x} \in \mathbb{R}^{MNL}$ is the vectorization of the SI, and ${\boldsymbol{n}\in \mathbb{R}^{{M}({N}+L-1)}}$ is additive noise from the acquisition process.

\textbf{Single Image Spatial-Spectral Image Super-Resolution:} A well-known task for SI is to recover a high spatio-spectral resolution SI from a low spatio-spectral resolution SI \cite{yan2018accurate}. In this task, the SI can be downsampled spatially and spectrally by a decimation matrix $\boldsymbol{D} \in \mathbb{R}^{\frac{MN}{k_s} \left(\frac{L}{k_l} \right) \times MNL}$, where $k_s$ and $k_l$ represent the decimation factor for the spatial and spectral resolution of the SI, respectively. The low spatio-spectral resolution image is represented as $\boldsymbol{y} = \boldsymbol{D} \boldsymbol{x} + \boldsymbol{n}$.

\textbf{SI Recovery from RGB Images:} Another task that has gathered significant attention from the research community is the mapping from RGB image to SI \cite{arad2022ntire}. This task consists on recover a SI from an RGB image with a count with less spectral information considering a known spectral response function $\boldsymbol{R} \in \mathbb{R}^{MN3 \times MNL}$. Then, the RGB image can be represented as $\boldsymbol{y} = \boldsymbol{R} \boldsymbol{x} + \boldsymbol{n}$.  

{Since the mentioned tasks are ill-posed problems, the objective is to recover the SI $\boldsymbol{x}$ from the observed data $\boldsymbol{y} = \boldsymbol{A}\boldsymbol{x}$,} where $\boldsymbol{A}$ could represent the sensing matrix $\boldsymbol{H}$, the decimation matrix $\boldsymbol{D}$ or the spectral response function $\boldsymbol{R}$, according to the selected problem. Then, we can solve the DL-based SI computational tasks through the optimization problem 
\begin{equation}
    {\boldsymbol{\hat{\theta}}} = \argmin_{\boldsymbol{{\theta}}} \frac{1}{K} \sum_{k=1}^{K} \Vert \boldsymbol{x}^{(k)} - \mathcal{M}_{\boldsymbol\theta}(\boldsymbol{y}^{(k)}) \Vert_2^2,
\end{equation}
{These are ill-posed inverse problems that are challenging. Therefore, we aim to increase the performance of the network $\mathcal{M}_{\theta}$ for each case by adding synthetic samples generated by the LD-GAN.}

\begin{figure*}[!t]
    \includegraphics[width=1\textwidth]{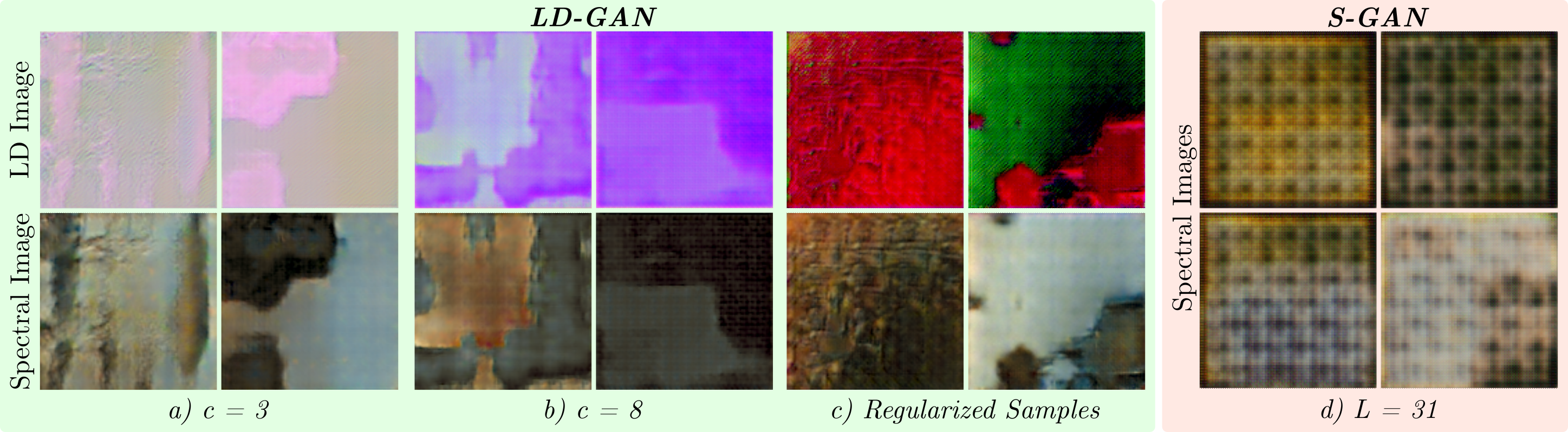}
    \centering
    \caption{Generated {images} for an LD representation with a) 3 bands and b) 8 bands. c) Generated SI samples from LD representation employing the statistical regularizer with $\mu_{gan} = \mu_{ae} = 0.001$.  For a-c) the top images refer to the generated LD images and the bottom images are the obtained SI by decoding the LD representation. d) Generated SIs without an LD representation. {The bottom row represents a false RGB mapping of generated SIs from their LD representations, except for S-GAN, which SIs is directly generated by the GAN.}}
    \vspace{-5mm}
    \label{fig:embedded-gan}
\end{figure*}

\section{Numerical Experiments}
\label{sec:results}
In this section, we perform several experiments with the AE network with different channels for the LD representation. For the GAN architecture, the deep convolutional generative adversarial network \cite{radford2015unsupervised} was adapted to the spatial size of the employed dataset, and the experiments are performed with respect to the LD image dataset obtained from the AE network against the entire-sized SI dataset. All experiments were performed on GPU with an NVIDIA RTX 3090 graphic card. The dataset employed for all the experiments is the ARAD 1K \cite{arad2022ntire}, which was preprocessed, reducing the spatial resolution to $256 \times 256$ and keeping the $31$ spectral bands. {This dataset contains $900$ samples for training and $50$ samples for testing.} We extract unique patches with a spatial resolution of $128 \times 128$ obtaining a total of 3600 patches for training and 200 patches for testing. {Recovery performance is measured with the peak-noise-to-signal ratio (PSNR) and the structural similarity index measure (SSIM) \cite{hore2010image}.}

\subsection{AE Experiments}

The compression of the SIs is only performed in the spectral dimension, where several values for the number of output channels $c = \{1, 3, 8, 16 \}$ of the encoder subnetwork are evaluated. {The encoder and subnetwork consist of 7 convolutional layers with ReLU activations, which decrease the number of {feature channels} from $16c$ to $c$ and the decoder network has also 7 layers increasing the number of features from $c$ to $16c$ and a final convolutional layer with $L=31$ features to match the SI spectral dimension}. Table \ref{tab:ae_results} shows that for the evaluated number of channels, $c=3$ has the best performance. Thus, this level of compression will be highly considered for the performance of the SI applications when the proposed method is applied.


\begin{table}[!ht]
\centering
\resizebox{0.5\linewidth}{!}{\begin{tabular}{c|c|c}\toprule[2pt]
 Channels & PSNR [dB] $\uparrow$  &  SSIM $\uparrow$  \\\toprule[2pt]
        1  & 33.72  & 0.9513  \\
        \cellcolor{green!12}3  & \cellcolor{green!12}39.32  & \cellcolor{green!12}0.9739  \\
        \cellcolor{blue!12}8  & \cellcolor{blue!12}38.37  & \cellcolor{blue!12}0.9722  \\
        16 & 38.31  & 0.9679  \\\toprule[2pt]
\end{tabular}} \vspace{-1em}
\caption{{AE recovery performance for different latent channel sizes. The highlighted green values are the best performance and the blue ones are the second best.}} \vspace{-1em}
\label{tab:ae_results}
\end{table}

\subsection{GAN Experiments}

\begin{figure}[!b]
    \centering
    \includegraphics[width=0.9\linewidth]{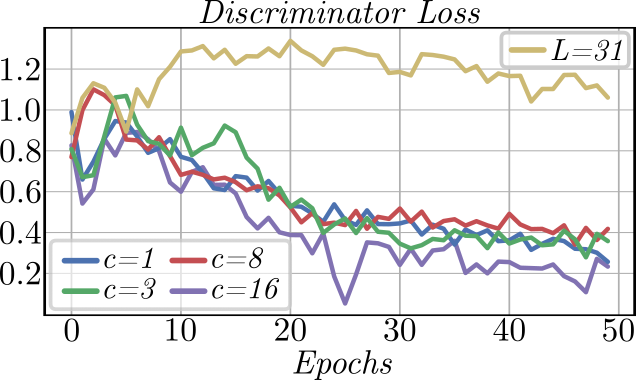}\vspace{-0.3cm}
    \caption{Discriminator loss function convergence for different channels. $L=31$ represents {the performance} by S-GAN.}
    \label{fig:loss}
\end{figure}

We employ the LD {images from} the SI dataset using the AE above mentioned to optimize the GAN. All GAN experiments were trained for 50 epochs with a learning rate of $2e^{-4}$ for both the discriminator and the generator. The generator and the discriminator networks employ 2D transpose convolution and 2D convolution layers {, respectively. Each subnetwork counts with batch normalization, and Leaky ReLU activations, after the convolutional layer, following the implementation given by \cite{radford2015unsupervised}}. All the experiments with GANs employed a random input $\boldsymbol{z} \in \mathbb{R}^{100} \sim \mathcal{N}(\boldsymbol{0}, \boldsymbol{1})$. In Fig.~\ref{fig:embedded-gan}, some of the generated LD samples and their respective SIs are shown for the LD-GAN with $c=\{3,8 \}$, LD-GAN with the statistical regularization, and S-GAN. To compare the effectiveness of the proposed method, we adopt the employed GAN for directly generating SIs without an LD representation, denoted as \textbf{S-GAN}. In Fig.~\ref{fig:loss}, the discriminator loss function is shown for different {channels} $c$. The performance shows this behavior properly when the GAN is trained on the LD {image dataset}. {This performance is far from the theoretical optimum when training is done with S-GAN.} 

\begin{figure*}[!t]
    \centering
    \includegraphics[width=1.0\linewidth]{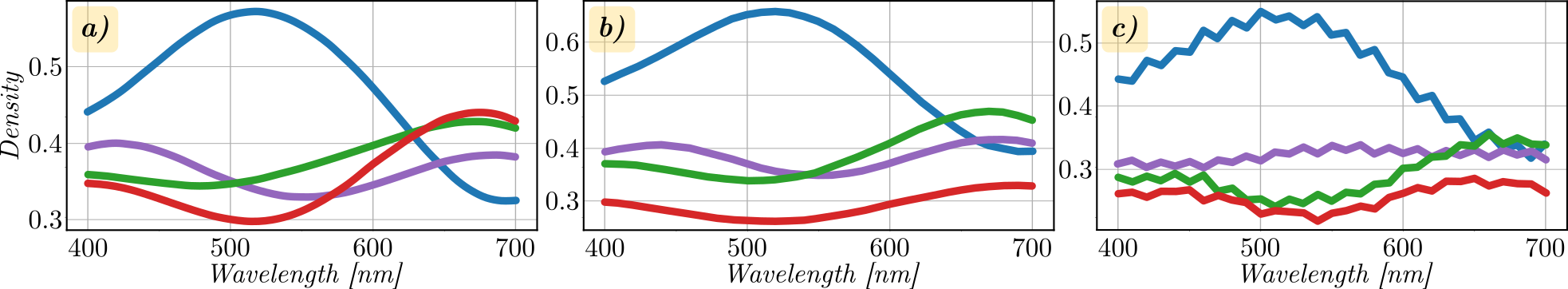}
    \caption{{Linear mixture model analysis of the generated images. Extracted $4$ endmembers from 100 SIs of the original SI dataset (a) the generated SI dataset with the proposed LD-GAN (b) and with the S-GAN (c).}}
    \label{fig:end}
\end{figure*}
\begin{figure*}[!t]
    \centering
    \includegraphics[width=1.0\linewidth]{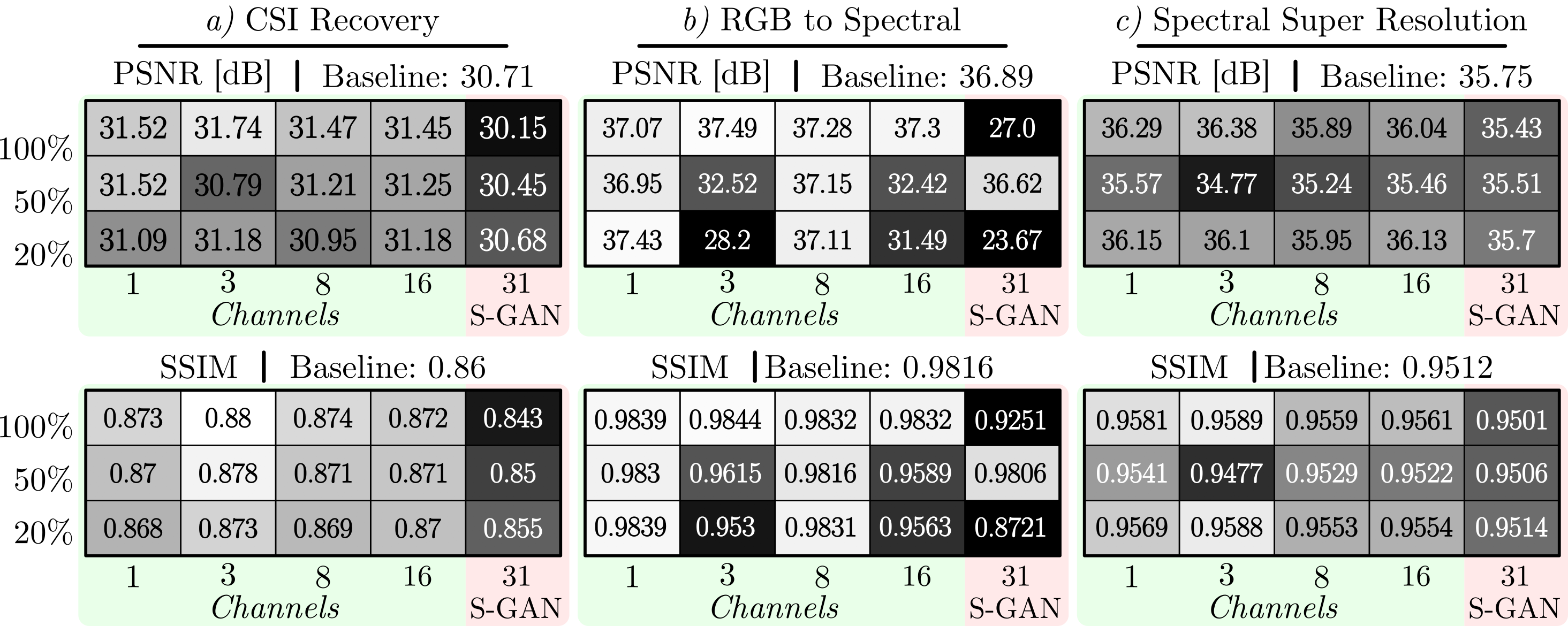}
    \caption{SI recovery performance in terms of PSNR and SSIM {employing} the proposed LD-GAN {considering the number of channels} and compared with the S-GAN for a different number of generated {SIs} {based on the percentage of samples from the original SI dataset} as a DA strategy for SI applications: (a) CSI recovery, (b) {RGB to spectral recovery}, and (c) SI super-resolution.}
    \vspace{-5mm}
    \label{fig:num_data}
\end{figure*}
We analyzed the generated spectral information with the proposed LD-GAN compared with the S-GAN. To this end, we employ a widely used SI analysis technique which is the linear mixture model \cite{eches2010bayesian}. This model states that a spectral signature of a given pixel of the SI is represented by the linear combination of the $q$ most representative signatures of the SI. These signatures are denoted as endmembers and the coefficients of the linear model are named abundance. To extract the endmembers and the abundances of a SI{, we employ} the hyperspectral vertex component analysis (HyperVCA) \cite{nascimento2005vertex} {algorithm}. {Specifically}, we employ HyperVCA to extract the $q=4$ endmembers from the original ARAD dataset, from the generated dataset by the LD-GAN and by the S-GAN, as shown in Fig{.}~\ref{fig:end} with the mean of 100 images. The obtained results illustrate that generated mean endmembers of the proposed LD-GAN  are very similar with respect to the original dataset. This result allows concluding that the generated spectral data follow similar behavior to the original dataset. While the S-GAN has many artifacts along the spectrum, which could affect negatively the training of networks a DA strategy for DL-based SI applications.


\begin{figure*}[!t]
    \includegraphics[width=1.0\linewidth]{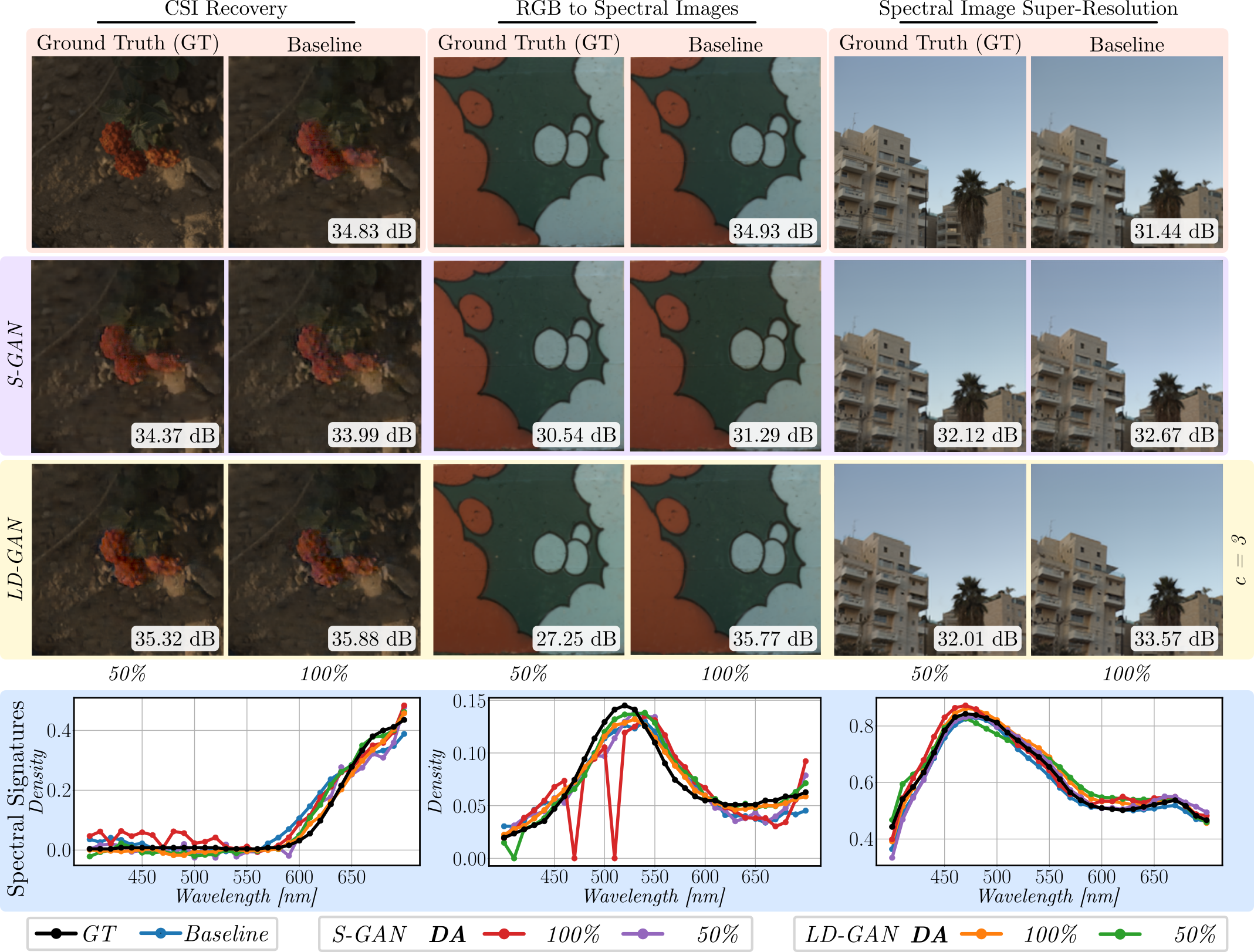}
    \caption{DL-based SI applications. Top to bottom: Ground Truth and baseline, S-GAN, and LD-GAN from $3$ bands LD image generation. From left to right: SI application for baseline (no DA based on GANs), $50\%$, and $100\%$ of DA based on GANs. Last row: Spectral signature from the middle coordinate for each application.}
    \vspace{-5mm}
    \label{fig:csi-results}
\end{figure*}


\begin{figure}[!t]
    \centering
    \includegraphics[width=1.0\linewidth]{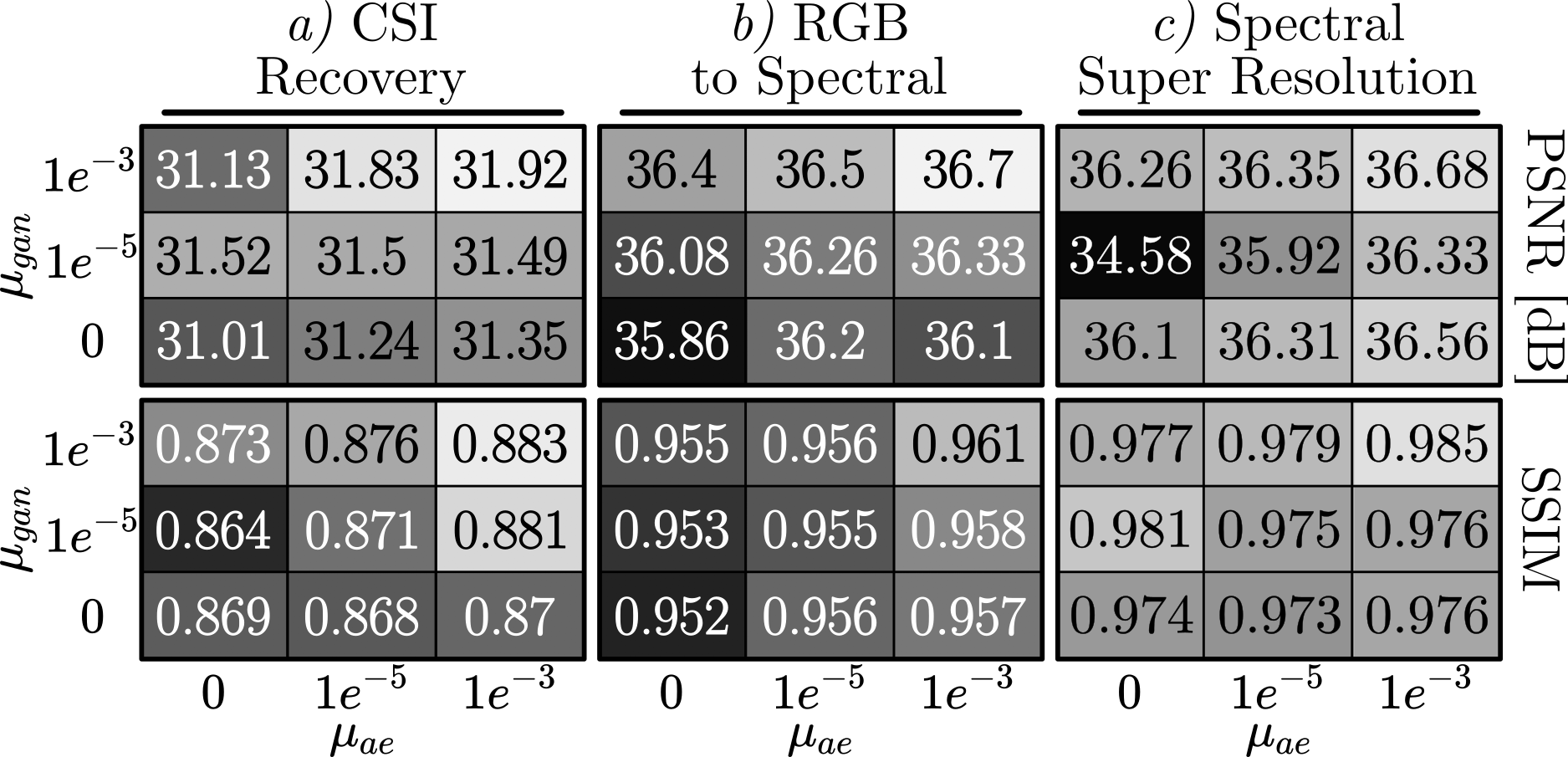}\vspace{-0.2cm}
    \caption{SI recovery performance in terms of PSNR and SSIM {employing} the proposed LD-GAN {with} different values of the regularization parameters $\mu_{gan}$ and $\mu_{ae}$ for each SI application.}
    \vspace{-5mm}
    \label{fig:reg}
\end{figure}

\subsection{DA Applications}

We validate the proposed method by generating a synthetic SI dataset and employing it as a DA for DL-based, CSI, SSSR, and RGB to spectral tasks. All of the following networks were trained for 100 epochs with a learning rate of $1e^{-3}$. From the obtained results, we report the best performance from testing images for each experiment.
\begin{itemize}
    \item \textit{CSI recovery setting:} The unrolled network deep spatial-spectral prior \cite{wang2019hyperspectral} was used for CSI recovery. The measurements were simulated through the CASSI system using a random CA with 50\% of transmittance.
    \item \textit{SI recovery from RGB images Setting:} A variation of the UNET architecture described for the SSSR case is employed. UNET-based models have been previously proposed for RGB to spectral task \cite{yan2018accurate}. To obtain the RGB input image for the network training and validation, the spectral response provided by the ARAD dataset \cite{arad2022ntire} {was used}. For the training of this network, the learning rate decays exponentially to achieve a more stable convergence.
    \item \textit{SI super-resolution setting:} We employed a UNET network with 4 downsampling and 4 upsampling convolutional blocks to reconstruct the high-resolution SI from low-spatial low-spectral measurements. The selection of the type of architecture is based on recent SSSR works \cite{wang2022fsl}. The spatial and spectral decimator factors are set as $k_s=2, k_l=4$.
    
\end{itemize}

\subsection{Quantitative Results}
For each SI application, we perform the recovery task considering each of the generated SI datasets by LD-GAN and S-GAN with $20\%$, $50\%$, and $100\%$ new samples, (percentages based on the number of samples from the original dataset). As shown in Fig.~\ref{fig:num_data}, the higher performances in terms of PSNR are achieved by LD-GAN with $c=3$. Furthermore, when a comparison between S-GAN and LD-GAN is performed, we can determine that not only LD-GAN is better, but S-GAN has a lower performance than the baseline {in several cases}. We also observe that the greater the number of data generated employed as DA the higher improvement of the performance for the SI applications is obtained with the proposed approach.

\subsection{Qualitative Results}
Visual results of a reconstructed SI from the test dataset are shown in Fig.~\ref{fig:csi-results}, where a high number of new samples allows a higher performance in terms of PSNR. The spectral signature extracted from the ground truth SI and its reconstructions shows that the training with the LD SI images achieves a density more similar to the ground truth data. The results suggest that the models trained with the synthetic samples from the proposed dataset have higher reconstruction quality in each task. 

\subsection{{Statistical Regularization Experiments}}
To validate the effectiveness of the proposed regularized training in the AE and GAN, we perform a hyperparameter study of the regularization parameters $\mu_{gan}$ and $\mu_{ae}$. The parameters $\mu_{ae} = \mu_{gan}  = \{0,1e^{-5},1e^{-3}\}$  were changed for each task. The augmented SI dataset was 100\% of the original dataset. Fig.~\ref{fig:reg} shows the performance of the mentioned experiment. The highest performance in each task is  obtained at higher regularization parameters showing the effectiveness of this training method. 

\begin{figure}[!t]
    \centering
    \includegraphics[width=\linewidth]{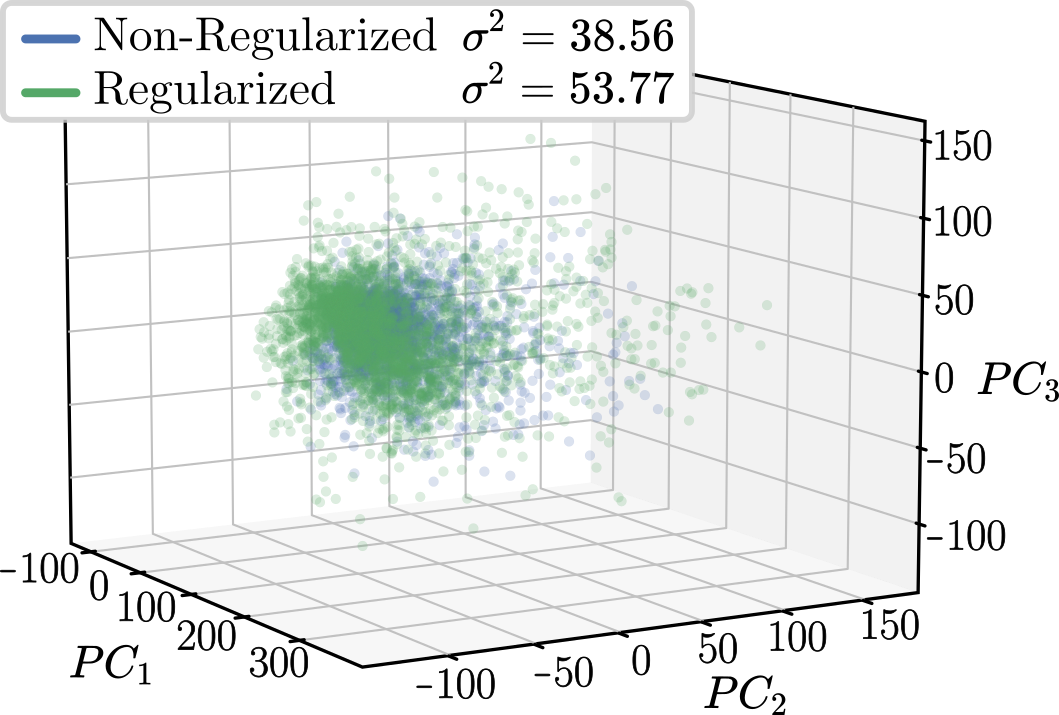}\vspace{-0.2cm}
    \caption{{Three principal components of the generated dataset and its respective variance with the regularized training and the non-regularized. $PC_1$, $PC_2$, and $PC_3$ refer to the first, second, and third principal components of the dataset respectively.}}
    \vspace{-5mm}
    \label{fig:dist}
\end{figure}

To visualize the effect of maximizing the variance of the generated LD representation in the GAN training, in Fig.~\ref{fig:dist}, the three first principal components of 3000 synthesized SIs were computed for the non-regularized GAN and the proposed LD-GAN with $\mu_{ae} = \mu_{gan} = 0.001$ showing that the last one has more variability than the first one. 
\subsection{Comparison with traditional DA techniques}
We compared the performance of the proposed method against the traditional DA consisting of rotations (rot), and vertical and horizontal flip (VFlip and HFlip), respectively, obtaining the same number of samples for each experiment reported in Table \ref{tb:csi_da_results}. The results obtained show the efficacy of the proposed method compared with the S-GAN and only geometric DA approaches. Additionally, the performance with the geometric DA achieves similar performance to the same experiments without this DA strategy, indicating that is not always helpful to use geometric DA.


\section{Conclusion and Future Work}
\label{sec:conclusion}
We proposed LD-GAN, a method that allows the generation of SIs through a GAN. To deal with the challenging issue of synthesizing the high-dimensional SI, we propose to train the GAN with an LD representation of a SI dataset obtained via the latent space of  an AE. Then the generative model synthesizes new LD samples mapped to the high-dimensional spectral image space with a decoder network. This approach significantly improves the convergence of the GAN. Moreover, a regularization based on data variability was proposed to optimally train the AE and GAN. We validate the proposed SI generation model as a DA strategy for some DL-based SI applications, such as CSI recovery, SISR, and SI recovery from RGB images, where an improvement is achieved from 0.5 to 1 dB. As future studies, improved performance of the LD-GAN can be obtained via a conditional generation\cite{isola2017image}, using, for instance, RGB images to guide the generation process. Training GAN with LD representation of the data can also be applied to other types of high-dimensional data such as video, polarized images, and 3D images for medical applications. 

\begin{table}[!t]
\resizebox{1\linewidth}{!}{\begin{tabular}{c|c|cc|cc|cc}\toprule[2pt]
 \multirow{2}{*}{Method} & \multirow{2}{*}{Geometric DA} &  \multicolumn{2}{c|}{CSI} & \multicolumn{2}{c|}{RGB to spectral} & \multicolumn{2}{c}{SISR}  \\
                          &         & PSNR [dB]        & SSIM   & PSNR [dB]      & SSIM         & PSNR [dB]      & SSIM    
                          \\\toprule[2pt]
                         \multirow{2}{*}{Baseline}     & \textcolor{red}{X}               & 29.75        & 0.827   & 37.01       & 0.982         & \cellcolor{blue!12}36.14      & 0.957   \\                                   & \checkmark              & 30.96        & 0.858   & 36.89       & 0.981         & 35.13      & 0.951   
                    \\\toprule[2pt]
                         \multirow{2}{*}{S-GAN}     & \textcolor{red}{X}               & 30.46        & 0.851   &    36.98    & \cellcolor{blue!12}0.981         & 35.72      & 0.954   \\                                   & \checkmark              & 30.15        & 0.843   & 27.00       & 0.925         & 35.43      & 0.950   
                    \\\toprule[2pt]
                         \multirow{2}{*}{LD-GAN}     & \textcolor{red}{X}               & 31.51 \cellcolor{blue!12}       & 0.874\cellcolor{blue!12}   & 37.38   \cellcolor{blue!12}    & 0.978         & 36.10      & \cellcolor{blue!12}0.957   \\                                   & \checkmark              & \cellcolor{green!12}31.74        & \cellcolor{green!12}0.880    & \cellcolor{green!12}37.49       &\cellcolor{green!12} 0.984         &\cellcolor{green!12} 36.38      & \cellcolor{green!12}0.958   
                    \\\toprule[2pt]
\end{tabular}} \vspace{-1em}
\caption{Comparison of performance for the SI computational tasks of the models trained with geometric DA and the GAN-based methods. The highlighted green values are the best performance and the blue ones are the second best} \vspace{-1em}
\label{tb:csi_da_results}
\end{table}
{Additionally, VAE can also be included instead of the AE, where the LD dataset employed to train the generative network is obtained from sampling the latent distribution of the VAE.}
{ Also, beyond GAN-based models, our method can be extended to state-of-the-art approaches based on diffusion probabilistic models (DPM). Some DPM generative approach employ LD representation from a VAE to condition the generation towards meaningful latent representation  \cite{Preechakul_2022_CVPR}. Different from this approach, the extension of our method to DPM aims to perform the corrupting process and the denoising of the inverse process over the LD representations of the dataset, and then, decode the generated image to obtain the desired synthetic data.
}

\bibliographystyle{ieee_fullname}
\bibliography{egbib}

\end{document}